\documentclass[10pt,twocolumn,letterpaper]{article}

\usepackage{iccv}
\usepackage{times}
\usepackage{epsfig}
\usepackage{graphicx}
\usepackage{amsmath}
\usepackage{amssymb}
\usepackage{booktabs}
\usepackage{multirow}
\usepackage{algorithm}
\usepackage{algorithmic}
\usepackage{xcolor}

\usepackage[pagebackref=true,breaklinks=true,letterpaper=true,colorlinks,bookmarks=false]{hyperref}

\usepackage{amsmath}
\usepackage[misc]{ifsym}
\DeclareMathOperator*{\argmax}{argmax}

\iccvfinalcopy

\begin{document}
\title{Self-training solutions for the ICCV 2023 GeoNet Challenge
}

\author{Team Name: CASIA-TIM (Lijun Sheng, Zhengbo Wang, and Jian Liang $^{\textrm{\Letter}}$) \\ 
Affiliation: Institute of Automation, Chinese Academy of Sciences (CASIA)\\
Contact author: Jian Liang (\textcolor{blue}{liangjian92@gmail.com})\\[4pt]
{\href{https://geonet-challenge.github.io/ICCV2023/challenge.html}{Challenge Website} \qquad
\href{https://eval.ai/web/challenges/challenge-page/2111/leaderboard}{Challenge Leaderboard}}
}

\maketitle
\ificcvfinal\thispagestyle{empty}\fi

\begin{abstract}
GeoNet is a recently proposed domain adaptation benchmark consisting of three challenges (i.e., GeoUniDA, GeoImNet, and GeoPlaces).
Each challenge contains images collected from the USA and Asia where there are huge geographical gaps. 
Our solution adopts a two-stage source-free domain adaptation framework with a Swin Transformer backbone to achieve knowledge transfer from the USA (source) domain to Asia (target) domain.
In the first stage, we train a source model using labeled source data with a re-sampling strategy and two types of cross-entropy loss.
In the second stage, we generate pseudo labels for unlabeled target data to fine-tune the model. 
Our method achieves an H-score of 74.56\% and ultimately ranks 1st in the GeoUniDA challenge.
In GeoImNet and GeoPlaces challenges, our solution also reaches a top-3 accuracy of 64.46\% and 51.23\% respectively.
The code is available at \href{https://github.com/tim-learn/GeoNet23_casia_tim}{Github}.
% The key aspects are shown below.
% \begin{enumerate}
% \item Team name, primary contact/author, and list of all other participants.

% \textcolor{blue}{CASIA-TIM (Lijun Sheng, Zhengbo Wang, and Jian Liang) (Contact author: liangjian92@gmail.com)}

% \item Key highlights/salient aspects of your approach.

% \textcolor{blue}{source-free domain adaptation, self-training strategy.}

% \item Total size and kinds of data used in both pre-training, fine-tuning and training phases.

% \textcolor{blue}{We use ImageNet-1k pre-trained model. The USA train split is used in source training and the Asia train split w/o labels is used in target adaptation.}

% \item Total model sizes (\# of parameters, type of architecture) and training strategies.

% \textcolor{blue}{Swin-B (88M parameters). Fine-tuning on source domain \& self-training on unlabeled target domain.}

% \item Complete list of foundational models, if any, used in training your algorithm.

% \textcolor{blue}{N/A. Only ImageNet-1k pre-trained Swin-B is used.}
% \end{enumerate}
\end{abstract}

%%%%%%%%% BODY TEXT
\section{Background}

\textbf{Source-free domain adaptation (SFDA)} \cite{liang2020we} is a popular test-time adaptation paradigm \cite{liang2023comprehensive} to achieve domain adaptation with only access to pre-trained source models and unlabeled target data.
Due to its privacy protection of source data and competitive performance compared to conventional domain adaptation, SFDA gains more attention in the transfer learning tasks with distribution shift.
SFDA methods always contain a source training stage and a target adaptation stage.
They fine-tune the pre-trained source model with only unlabeled target data through unsupervised learning strategies, such as pseudo-labeling, consistency, and clustering.
Note that SFDA methods are not allowed to access the source dataset during adaptation.

\textbf{Swin Transformer} \cite{liu2021swin} is a strong backbone for a broad range of computer vision tasks.
Its efficiency and flexibility benefit from the design of shifted windows and the hierarchical architecture.
Especially, Swin Transformer base model (Swin-B) obtains an accuracy of 86.4\% on ImageNet-1K benchmark.
Due to its competitive performance on ImageNet-1K benchmark and attractive generalization ability, we choose Swin-B as our backbone network in all experiments.

\section{Framework}
\subsection{Overview}
We adopt a two-stage source-free domain adaptation framework \cite{liang2020we} for all three challenges in GeoNet benchmark.
Our solution consists of a source training stage and a target adaptation stage.
In the source training stage, we pre-train a source model using labeled USA images with the initialization of ImageNet-1k pretrained Swin-B model.
In the target adaptation stage, pseudo labels of unlabeled Asia images are introduced and the source model is fine-tuned through the self-training strategy to improve its performance in Asia domain.
Due to the source training stages being the same for all challenges, we describe its process in the next subsection.
Target adaptation details for the three challenges are provided in their respective sections.

\subsection{\textcolor{blue}{Source Training}} 
\label{source training}
Inspired by SHOT \cite{liang2020we}, our model $f(x)$ consists of a feature extractor $g: \mathcal{X}\to \mathbb{R}^{d}$ and a classifier module $h: \mathbb{R}^{d}\to \mathbb{R}^{K}$, i.e., $f(x) = h\left(g(x)\right)$.
$d$ is the dimension of feature space which is set to 256 and $K$ is the number of categories.
Feature extractor $g$ consists of a Swin-B backbone, a linear layer for dimension transformation, and a batch normalization (BN) layer.
The classifier module $h$ is designed as a standard linear layer whose output dimension equals the category number.

In the source training stage, we are given source training dataset $\{x_s^{i},y_s^{i}\}_{i=1}^{n_s}$ from USA domain $\mathcal{D}_s$ to train a source model $f_s(x)=h_s(g_s(x))$.
We adopt cross-entropy loss with the label smoothing technique for better generalization ability.
The objective function is given by,
\begin{equation}
    \begin{aligned}
        \mathcal{L}_{src}^{ls} = -\mathbb{E}_{(x_s,y_s)} \sum\nolimits_{k=1}^{K} q_k^{ls} \log \delta_k(f_s(x_s)),
    \end{aligned}
\end{equation} 
where $q^{ls}_k=(1-\alpha)q_k + \alpha/K$ is the smoothed label and $\alpha$ is set to 0.1, and $\delta$ denotes soft-max operation.

Also, we use the exponential moving average (EMA) mechanism to maintain an EMA model that has better intermediate representations.
EMA model parameters are updated once per epoch using an EMA mechanism with a smoothing coefficient of 0.95.
We use cross-entropy loss between the model prediction and the EMA model prediction, which is given as,
\begin{equation}
    \begin{aligned}
        \mathcal{L}_{src}^{EMA} = -\mathbb{E}_{(x_s,y_s)} \sum\nolimits_{k=1}^{K} \delta_k(f_s^{EMA}(x_s)) \log \delta_k(f_s(x_s)).
    \end{aligned}
\end{equation} 

Considering that the category distribution of the source domain may be inconsistent with the target, we introduce a re-sampling strategy to mitigate the negative impact of the potential inconsistency.
For the $c$-th class, the weight $\mathcal{W}_{c}$ is calculated by the reciprocal of the proportion of the category sample number ${N}_{i}$ to the total:
\begin{equation}
    \begin{aligned}
        \mathcal{W}_{c} = \frac{\sum\nolimits_{i=1}^{K}\mathcal{N}_{i}}{\mathcal{N}_{c}}.
    \end{aligned}
\end{equation}

To summarize, during the source training stage, a re-sampling strategy is introduced and the full objective is as,
\begin{equation}
    \begin{aligned}
        \mathcal{L}_{src} = \mathcal{L}_{src}^{ls} + \lambda_{EMA} \mathcal{L}_{src}^{EMA},
    \end{aligned}
\end{equation} 
where $\lambda_{EMA}$ stays at zero for the first 40\% of iterations, and then switches to one for the rest of the training process.

\section{Challenge track: UniDA on GeoUniDA}
\subsection{Task}
GeoUniDA \cite{kalluri2023geonet} is a universal domain adaptation challenge with a huge geographical gap.
Both source and target domains have 62 shared (a.k.a., known) classes and 138 private (a.k.a., unknown) classes.
The goal of the task is to classify the samples from the shared classes correctly and mark the samples from the target private classes as ``unknown".
The performance is evaluated by the H-score metric which is the harmonic mean of the known accuracy and the binary unknown accuracy.

\subsection{\textcolor{blue}{Target Adaptation}} \label{unida_adaptation}
After employing the source training process in Sec.~\ref{source training}, we fine-tune the source model with unlabeled target dataset $\{x_t^{i}\}_{i=1}^{n_t}$.
Following SHOT \cite{liang2020we}, we use the source model as initialization, freeze the classifier module $h_t$, and optimize the feature extractor $g_t$.

We divide the target domain samples into known set $\mathcal{S}_k$ and unknown set $\mathcal{S}_u$ for deploying different optimization strategies.
Two thresholds (i.e., $\tau_{high}$ and $\tau_{low}$) are introduced to define the boundary of two sets.
$\mathcal{S}_k$ consists of samples whose entropy of prediction is lower than $\tau_{low}$.
Samples with entropy greater than $\tau_{high}$ belong to $\mathcal{S}_u$:
\begin{equation}
    \begin{aligned}
        \mathcal{S}_k &= \{ x_t \| \mathcal{H}(x_t) \leq \tau_{low} \}, \\
        \mathcal{S}_u &= \{ x_t \| \mathcal{H}(x_t) \geq \tau_{high} \},
    \end{aligned}
\end{equation}
where $\mathcal{H}(x_t)=-\sum\nolimits_{k=1}^{K} \delta_k(h_t(g_t(x_t))) \log \delta_k(h_t(g_t(x_t)))$ denotes the entropy of target data $x_t$.

For stable training, we update $\tau_{high}$ and $\tau_{low}$ from 
initial value 0.5 consecutively:
\begin{equation}
    \begin{aligned}
        \tau_{high} &= 0.5 + 0.2 \times \zeta, \\
        \tau_{low} &= 0.5 - 0.2 \times \zeta,
    \end{aligned}
\end{equation}
where $\zeta\in(0,1)$ is a variable that increases uniformly.

For samples in $\mathcal{S}_k$, we treat them as known samples and calculate cross-entropy loss between their predictions and pseudo labels $\hat{y}_t$:
\begin{equation}
\begin{aligned}
    \mathcal{L}_k(g_t) &= - \mathbb{E}_{(x_t,\hat{y}_t)\in \mathcal{S}_k \times \hat{\mathcal{Y}}_t } \log \delta_{\hat{y}}(h_t(g_t(x_t))),\\
    \hat{y}_t &= \argmax_k \ \delta_k(h_t(g_t(x_t)
    )).
\end{aligned}
\end{equation}
For samples in $\mathcal{S}_u$, we treat them as unknown samples and try to maximize their entropy to improve the outlier recognition ability:
\begin{equation}
\begin{aligned}
    \mathcal{L}_u(g_t) = -\mathbb{E}_{x_t\in \mathcal{S}_u} & \mathcal{H}(x_t).\\
\end{aligned}
\end{equation}
To summarize, during the target adaptation stage, the full objective is as,
\begin{equation}
    \begin{aligned}
        \mathcal{L}_{tar}^{UniDA}(g_t) = \mathcal{L}_k(g_t) + \alpha \mathcal{L}_u(g_t),
    \end{aligned}
\end{equation} 
where $\alpha$ is a trade-off hyperparameter which is
set to 0.3.

\subsection{Result}
We evaluate our solution on Asia test split and challenge test set of GeoUniDA challenge and results are shown in Table~\ref{tab:unida}.
We use the last source checkpoint as the source model in `old' methods and use the best one in `new' methods. 
Besides, `source' refers to the performance of the source model, and `adapt' refers to the performance after target adaptation.
Based on the provided results, our solution achieves an H-score of 74.56\% on the challenge test set which is the 1st place in the leaderboard.

\setlength{\tabcolsep}{4.0pt}
    \begin{table}[!t]
        \centering
        \caption{H-score (\%) on Asia test split and challenge test set of GeoUniDA challenge (`old/new' refers to last/best source model checkpoint, `source/target' refers to performance before/after target adaptation stage).}
        \vspace{1mm}
        \resizebox{0.4\textwidth}{!}
        {
            \begin{tabular}{l|c|c}
            \toprule
            Method & Asia test split & Challenge (eval.ai) \\
            \midrule
            Baseline & - & 53.59 \\
            old source & 61.47 & 70.24 \\
            old adapt & 64.34 & 74.43 \\
            new source & 62.03 & 71.08 \\
            new adapt & 64.45 & \textbf{74.56} \\
            \bottomrule
            \end{tabular}
        }   
    \label{tab:unida}
    \end{table}

\section{Challenge track: UDA on GeoPlaces}
\subsection{Task}
GeoPlaces \cite{kalluri2023geonet} is an unsupervised domain adaptation challenge with a large huge geographical gap.
It consists of images from 204 places across two domains (i.e., USA and Asia).
The goal of the task is Asian place recognition by utilizing the labeled set from the USA and the unlabeled data from Asia.
The performance is evaluated by the top-3 accuracy.

\subsection{\textcolor{blue}{Target Adaptation}}
In the target adaptation stage, the model is fine-tuned with unlabeled target dataset $\{x_t^{i}\}_{i=1}^{n_t}$.
Following SHOT \cite{liang2020we}, we use the source model as initialization, freeze the classifier module $h_t$, and optimize the feature extractor $g_t$.

The self-training strategy calculates the cross-entropy loss between the model prediction and the pseudo label.
Due to the difficulty of GeoPlaces challenge, the accuracy of the pseudo label is too low
Thus We choose both the first and second highest categories as double pseudo labels to provide supervision signals.

The generation of two pseudo labels of $\{x_t^{i}\}_{i=1}^{n_t}$ is as,
\begin{equation}
    \begin{aligned}
        \hat{y}_t^{1st} &= \argmax_k \ \delta_k(f_s(x_t)), \\
        \hat{y}_t^{2nd} &= \argmax_{k \neq \hat{y}_t^{1st}} \ \delta_k(f_s(x_t)).
    \end{aligned}
\end{equation}
Based on double pseudo labels $\hat{y}_t^{1st}, \hat{y}_t^{2nd}$, the cross-entropy loss is calculated as,
\begin{equation}
    \begin{aligned}
        \mathcal{L}_{tar}^{Places}(g_t) &= - \beta \ \mathbb{E}_{(x_t,\hat{y}_t^{1st})\in \mathcal{X}_t \times \hat{\mathcal{Y}}_t }  \log \delta_{\hat{y}_t^{1st}}(h_t(g_t(x_t))) \\
        & -\gamma \ \mathbb{E}_{(x_t,\hat{y}_t^{2nd})\in \mathcal{X}_t \times \hat{\mathcal{Y}}_t }  \log \delta_{\hat{y}_t^{2nd}}(h_t(g_t(x_t))),
    \end{aligned}
\end{equation}
where $\beta$ and $\gamma$ are trade-off hyperparameters which are set to 0.3 and 0.1 respectively.

\subsection{Result}
We evaluate our solution on Asia test split and challenge test set of GeoPlaces challenge and results are shown in Table~\ref{tab:places}.
Our solution achieves a top-3 accuracy of 51.23\% on the challenge test set.

\setlength{\tabcolsep}{4.0pt}
    \begin{table}[!t]
        \centering
        \caption{Top-3 Accuracies (\%) on Asia test split and challenge test set of GeoPlaces challenge.}
        \vspace{1mm}
        \resizebox{0.4\textwidth}{!}
        {
            \begin{tabular}{l|c|c}
            \toprule
            Method & Asia test split & Challenge (eval.ai) \\
            \midrule
            Baseline & - & 41.60 \\
            old source & 66.36 & 50.24 \\
            old adapt & 67.73 & \textbf{51.23} \\
            new source & 66.52 & 50.56 \\
            new adapt & 67.26 & 50.85 \\
            \bottomrule
            \end{tabular}
        }   
    \label{tab:places}
    \end{table}

\section{Challenge track: UDA on GeoImNet}
\subsection{Task}

GeoImNet \cite{kalluri2023geonet} is a challenging object classification benchmark whose images are collected from two continents with large geographical gaps.
It contains two distant domains (i.e., USA and Asia) and 600 object categories.
The task aims to utilize the annotation knowledge in the USA domain to improve the classification performance in the unlabeled images from Asia domain.
The performance is evaluated by the top-3 accuracy.

\subsection{\textcolor{blue}{Target Adaptation}}
In the target adaptation stage, the model is fine-tuned with unlabeled target dataset $\{x_t^{i}\}_{i=1}^{n_t}$.
Following SHOT \cite{liang2020we}, we use the source model as initialization, freeze the classifier module $h_t$, and optimize the feature extractor $g_t$.

We employ the self-training strategy which learns representations under the supervision of the pseudo label.
To obtain higher quality pseudo-labels, inspired by SHOT \cite{liang2020we}, we attain the centroid for $k$-th category in the feature space,
\begin{equation}
    c_k = \frac{\sum_{x_t\in \mathcal{X}_t}{\delta_k(h_s(g_s(x_t)))}\ {g}_s(x_t)}{\sum_{x_t\in \mathcal{X}_t}{\delta_k(h_s(g_s(x_t)))}}.
\end{equation}
Then, the pseudo labels are obtained based on the cosine distance between samples and centroids,
\begin{equation}
    \hat{y}_t = \argmax_k\ cosine({g}_s(x_t), c_k).
\end{equation}
Based on the pseudo label $\hat{y}_t$, the cross-entropy loss is calculated as,
\begin{equation}
    \begin{aligned}
        \mathcal{L}_{tar}^{ImNet}(g_t) = - \eta \ \mathbb{E}_{(x_t,\hat{y}_t)\in \mathcal{X}_t \times \hat{\mathcal{Y}}_t } \log \delta_{\hat{y}_t}(h_t(g_t(x_t))),
    \end{aligned}
\end{equation}    
where $\eta$ is a hyper-parameter which is set to 0.3.

\subsection{Result}
We evaluate our solution on Asia test split and challenge test set of GeoImNet challenge and results are shown in Table~\ref{tab:imnet}.
Our solution achieves a top-3 accuracy of 64.49\% on the challenge test set.
\setlength{\tabcolsep}{4.0pt}
    \begin{table}[!t]
        \centering
        \caption{Top-3 Accuracies (\%) on Asia test split and challenge test set of GeoImNet challenge.}
        \vspace{1mm}
        \resizebox{0.4\textwidth}{!}
        {
            \begin{tabular}{l|c|c}
            \toprule
            Method & Asia test split & Challenge (eval.ai) \\
            \midrule
            Baseline & - & 49.65 \\
            old source & 69.83 & 63.31  \\
            old adapt & 70.61 &  \textbf{64.49} \\
            new source & 70.00 & 63.29 \\
            new adapt & 70.85 & 64.46 \\
            \bottomrule
            \end{tabular}
        }   
    \label{tab:imnet}
    \end{table}

\section{Implementation Details}

In this section, we provide the details of the whole process.
We use Swin Transformer Base (Swin-B)~\cite{liu2021swin} as the backbone in all experiments and we load ImageNet-1k pre-trained parameters provided by Torchvision.
A \{Linear-BN-Linear\} module is followed by Swin-B backbone and the total number of the parameters is 87,160,336 (less than 88M).

We adopt the learning rate scheduler $\eta=\eta_0 \cdot (1+10\cdot p)^{-1.0}$, where $p$ is the training progress changing from 0 to 1.
The initial learning rate $\eta_0$ is 1e$^{-3}$ for Swin-B backbone and 1e$^{-2}$ for the rest in both the source training and target adaptation stage.
We run source training for 10 epochs, while the maximum number of epochs during target adaptation for GeoUniDA, GeoPlaces, and GeoImNet is set to 5, 1, and 1, respectively. 
Besides, the batch size is set to 64 in all experiments.

\newpage
{\small
\bibliographystyle{ieee}
\bibliography{ref}
}

\end{document}